\documentclass[runningheads]{llncs}

\usepackage{eccv}

\usepackage{eccvabbrv}

\usepackage{graphicx}
\usepackage{booktabs}
\usepackage{multirow}
\usepackage[accsupp]{axessibility}  
\usepackage{wrapfig}

\usepackage{hyperref}

\usepackage{orcidlink}

\begin{document}

\title{DeCo: Decoupled Human-Centered Diffusion Video Editing with Motion Consistency} 

\titlerunning{DeCo}

\author{Xiaojing Zhong\inst{1,2} \and
Xinyi Huang\inst{1} \and
Xiaofeng Yang\inst{2} \and Guosheng Lin\inst{2}\thanks{Corresponding Authors} \and Qingyao Wu\inst{1,3,\footnotemark[1]}}

\authorrunning{X.Zhong et al.}

\institute{School of Software Engineering, South China University of Technology \\
 \and 
Nanyang Technological University\\ 
\and 
Peng Cheng Laboratory \\
\email{vzxj12@gmail.com, gslin@ntu.edu.sg, qyw@scut.edu.cn}}

\maketitle

\begin{abstract} Diffusion models usher a new era of video editing, flexibly manipulating the video contents with text prompts. Despite the widespread application demand in editing human-centered videos, these models face significant challenges in handling complex objects like humans. In this paper, we introduce DeCo, a novel video editing framework specifically designed to treat humans and the background as separate editable targets, ensuring global spatial-temporal consistency by maintaining the coherence of each individual component. Specifically, we propose a decoupled dynamic human representation that utilizes a parametric human body prior to generate tailored humans while preserving the consistent motions as the original video. In addition, we consider the background as a layered atlas to apply text-guided image editing approaches on it. To further enhance the geometry and texture of humans during the optimization, we extend the calculation of score distillation sampling into normal space and image space. Moreover, we tackle inconsistent lighting between the edited targets by leveraging a lighting-aware video harmonizer, a problem previously overlooked in decompose-edit-combine approaches. Extensive qualitative and numerical experiments demonstrate that DeCo outperforms prior video editing methods in human-centered videos, especially in longer videos.

\end{abstract}

\section{Introduction}
\label{sec:intro}


Text-driven video editing has achieved remarkable advances with the emergence of diffusion models \cite{ho2020denoising,sohl2015deep,song2019generative,yang2023lods,Yang_Liu_Xu_Su_Wu_Lin_2024}, which have garnered widespread acclaim for their user-friendly characteristics. Given the significant application value of human-centered videos across various fields, such as movie production, virtual try-on, and gaming, delving into editing these videos becomes paramount. With the strides achieved by diffusion models in text-guided image editing \cite{hertz2022prompt,parmar2023zero,mokady2023null,brooks2023instructpix2pix,tumanyan2023plug}, a natural solution would be to extend these approaches to the video domain. Nevertheless, propagating the edited content throughout the entire video still poses a formidable challenge. Beyond performing precise edits that adhere to text prompts, it's imperative to maintain a motion trajectory consistent with the original video. 


\begin{figure}[t]
\centering
\setlength{\abovecaptionskip}{0pt}
\includegraphics[width=1.0\textwidth,height=0.5\textheight]{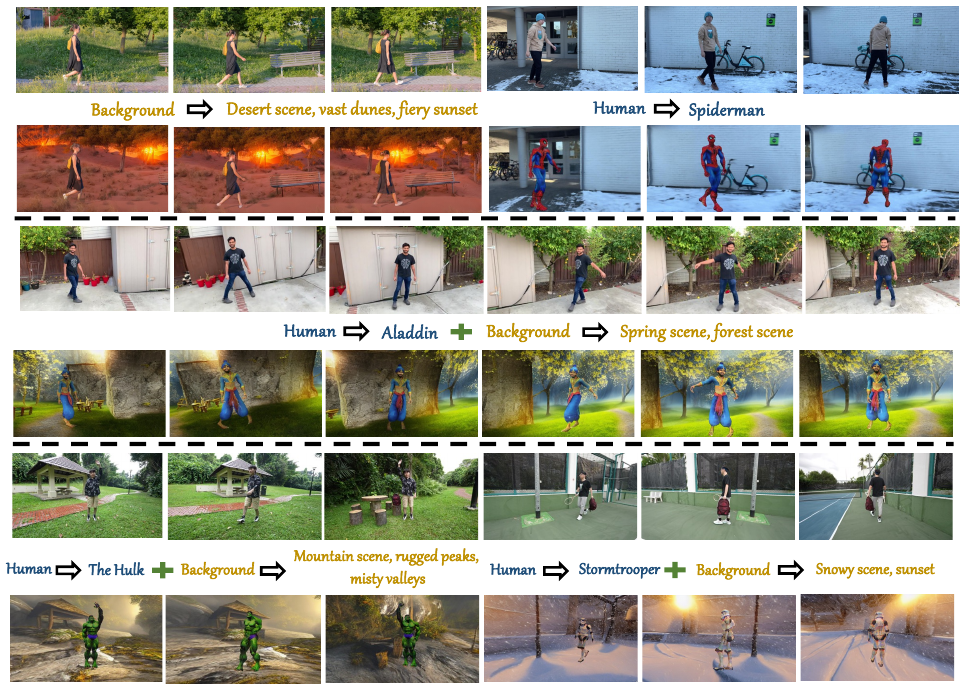}
\caption{\textbf{Editing results.} Given text prompts, our proposed \textbf{DeCo} enables independent and combined edits of both background and humans while preserving the original structure and motions. Best viewed in zoom.}
\label{1}
\end{figure}

To enhance frame-wise editing, pioneer Tune-A-Video \cite{wu2023tune} leverages Text-to-Image (T2I) diffusion models to edit the video through tuning the spatial-temporal attention module during the reconstruction of the original video. Following works \cite{ceylan2023pix2video,qi2023fatezero,liu2023video,yang2023rerender,liang2023flowvid} integrate condition maps (e.g., depth, edges) or optical flow into their frameworks to ensure temporal smoothness and control over the structure. However, they are both plagued by problems of appearance distortion and structural flickering as the number of frames increases. Several works \cite{ouyang2023codef,kasten2021layered} explore novel video representations with an objective: to transform videos into a unified image format. This transformation allows for the direct utilization of image-based processing algorithms across the entire video, circumventing the conventional method of processing it frame by frame. CoDeF \cite{ouyang2023codef} encodes video content within a canonical field, subsequently utilizing a deformation field to warp the canonical image to the video frames over the temporal dimension. CoDeF excels in adapting image algorithms for video processing. However, it struggles with local adjustments due to the global treatment of the canonical image. Neural Layered Atlases (NLA) \cite{kasten2021layered} maps each pixel of every frame in the video to a corresponding 2D uv map. Pixels are sampled based on their uv coordinates to create an atlas representing the editable object (or background). One can significantly maintain temporal consistency by editing the atlas and mapping it back to the video according to uv coordinates. Building on NLA, \cite{bar2022text2live,chai2023stablevideo,lee2023shape} implement flexible manipulation over the editable atlas, enhancing the control of edits. However, these approaches composite the edited objects onto the background without considering the need for harmonization between them. Furthermore, they often encounter challenges in dealing with complex objects such as humans, who exhibit intricate appearances and non-rigid movements within videos.

In this paper, we propose DeCo, an innovative video editing framework centred on human motion consistency, aiming to render the target human with diverse appearances and shapes within a text-specified scene. Instead of propagating key frames across the entire video, we consider the background and humans as separate targets for editing. Specifically, we employ depth-guided image algorithms to edit the background represented as a layered atlas for temporal consistency. To avoid spatial distortion between frames caused by the coupling of appearance and dynamic motions, we leverage an explicit human body prior, SMPL-X \cite{pavlakos2019expressive}, to drive the human with motion sequences. This harness the power of 3D human pose estimation methods \cite{li2021hybrik,cai2023smpler,wei2022capturing,yu2023gla,qiu2023psvt,zhong2021mv}, which can regress the pose parameters of SMPL-X from a video sequence to represent a minimally clothed human mesh. In order to tailor the mesh to conform to text prompts, we optimize it under the canonical pose with vertex-based displacements and a texture map using Score Distillation Sampling (SDS), where the denoising scores are computed not only in the latent space but also in the normal space and image space to enhance geometry and texture details. As prior methods \cite{chai2023stablevideo,bar2022text2live,lee2023shape} composite the rendered atlases directly without considering lighting consistency, we propose a video harmonizer to ensure harmonization of their albedo and shading maps, making the whole video appear realistic through relighting the objects. Fig. \ref{1} showcases results from both independent and combined edits. Overall, our contributions are summarized as follows:


\begin{itemize}
\item{ We design a human-centered, text-guided video editing framework that facilitates diverse editing of both scenes and humans. A decoupled dynamic human representation is proposed to ensure consistent motions based on a parametric human body prior, which 
avoids distortion induced by the coupling of appearance and complex motions. Additionally, we edit scenes within the layered atlas space to preserve temporal coherence.} 


\item{ We enhance the geometry and texture details of humans by extending the score distillation process into both normal and image space.} 
\item{ We propose a lighting-aware video harmonizer to ensure lighting consistency between the edited targets.}
\end{itemize}

\section{Related work}
\subsection{Diffusion-based Video Editing}

Recent advancements in extending diffusion-based image editing methods \cite{zhang2023adding,rombach2022high,brooks2023instructpix2pix,kawar2023imagic} to the video domain have seen notable progress. Works like Tune-A-Video \cite{wu2023tune} inflate spatial self-attention layers into spatial-temporal self-attention layers for enhanced temporal modelling, calculating sparse causal attention between initial and preceding frames. FateZero \cite{qi2023fatezero} merges self-attention and cross-attention feature maps from both DDIM inversion and denoising processes. Video-P2P \cite{liu2023video} extends Prompt-to-Prompt \cite{hertz2022prompt} to videos by optimizing a shared unconditional embedding for video inversion. TokenFlow \cite{geyer2023tokenflow} preserves the inter-frame correspondence and redundancies of the original video features within the edited features. Structural guidance, such as depth maps and optical flow, is increasingly integrated into frameworks for spatial constraints \cite{zhang2023controlvideo,ma2023follow}. Rerender A Video \cite{yang2023rerender} employs edge maps and optical flow to impose cross-frame shape constraints. However, temporal consistency remains a challenge, particularly in longer videos, due to the intrinsic deficiency in extending T2I models into the video domain.

 \par 
Several approaches propose novel video diffusion models or represent video inputs in different forms. Gen-1 \cite{esser2023structure} and VideoComposer \cite{wang2023videocomposer} develop video diffusion models, receiving different condition maps to steer the generation of motion and appearance in videos. CoDeF \cite{ouyang2023codef} decomposes videos into the canonical content field and temporal deformation field. These methods frequently face issues with separating objects from their backgrounds during the editing process, making it hard to achieve precise manipulation. Text2live \cite{bar2022text2live} separates editable objects from the background in videos using 2D neural layered atlases (NLA) \cite{kasten2021layered}. In NLA, positions in the video are mapped to their corresponding locations within each atlas layer, enabling targeted editing. \cite{lee2023shape} allows for shape editing that relies on dense semantic correspondence learning. StableVideo \cite{chai2023stablevideo} leverages depth cues to guide background edits and uses canny edge detection for object modification. While these NLA-based methods succeed in creating coherent video frames, they encounter challenges in editing complex objects like humans, often leading to distorted appearances. Moreover, these methods overlook the crucial aspect of harmonizing edited objects with their backgrounds, which is essential for maintaining visual consistency. 


\subsection{Text-driven Human Generation}

Avatar-CLIP \cite{hong2022avatarclip} initializes 3D human
geometry via a shape VAE network and harnesses the power of CLIP loss \cite{radford2021learning} to refine geometry and texture. DreamFusion \cite{poole2022dreamfusion} introduces the SDS loss and optimizes NeRF \cite{mildenhall2021nerf} under the strong image prior to the T2I diffusion model. However, Text-to-3D efforts \cite{chen2023fantasia3d,poole2022dreamfusion,li2023sweetdreamer,chen2023it3d,zhu2023hifa} struggle to produce high-fidelity human meshes when tasked with generating text-driven 3D humans. DreamAvatar \cite{cao2023dreamavatar} and AvatarCraft \cite{jiang2023avatarcraft} leverage the parameterized SMPL model as a shape prior, yet their ability to obtain complex human attributes such as clothing and hair remains limited. Addressing this, Dreamwaltz \cite{huang2023dreamwaltz} enhances the SDS loss with a 3D-aware skeleton conditioning, while Humannorm \cite{huang2023humannorm} and AvatarVerse \cite{zhang2023avatarverse} employ the hybrid 3D representation DMTet \cite{shen2021deep}, along with structural condition maps, for more detailed and realistic geometrical outputs. Notwithstanding the high quality of static human meshes these methods can create, the necessity of a rigged skeleton for animation introduces significant computational demands, particularly for dynamic video scenes. TADA \cite{liao2023tada} expands upon the upsampled SMPL-X model \cite{pavlakos2019expressive}, integrating a displacement layer and texture map to support the generation of diverse and animated characters. However, TADA blends the SDS loss from image and normal maps without constraints, leading to results that fall short of satisfaction.
\subsection{Image and Video Harmonization}
Image harmonization aims to blend foreground and background seamlessly within a composite image by aligning their illumination and colour statistics. Scs-co \cite{hang2022scs} incorporates contrastive learning into the harmonization. \cite{sofiiuk2021foreground} leverages semantic features derived from pre-trained models. \cite{tan2023deep} focus on the correlation and decorrelation within dual colour space. \cite{guo2021image} proposes a transformer-based architecture for image harmonization. Furthermore, \cite{careaga2023intrinsic,guo2021intrinsic} apply intrinsic image decomposition to image harmonization, separating the composite image into reflectance and illumination for targeted harmonization. Harmonizing composite frames individually using image harmonization techniques can lead to flickers and artifacts. \cite{huang2019temporally} and \cite{lu2022deep} utilize optical flow and colour mapping consistency, respectively, to ensure consistency between adjacent frames throughout the harmonized video sequence.
\begin{figure}[t]
\centering
\setlength{\abovecaptionskip}{0pt}
\includegraphics[width=1.0\textwidth,height=0.36\textheight]{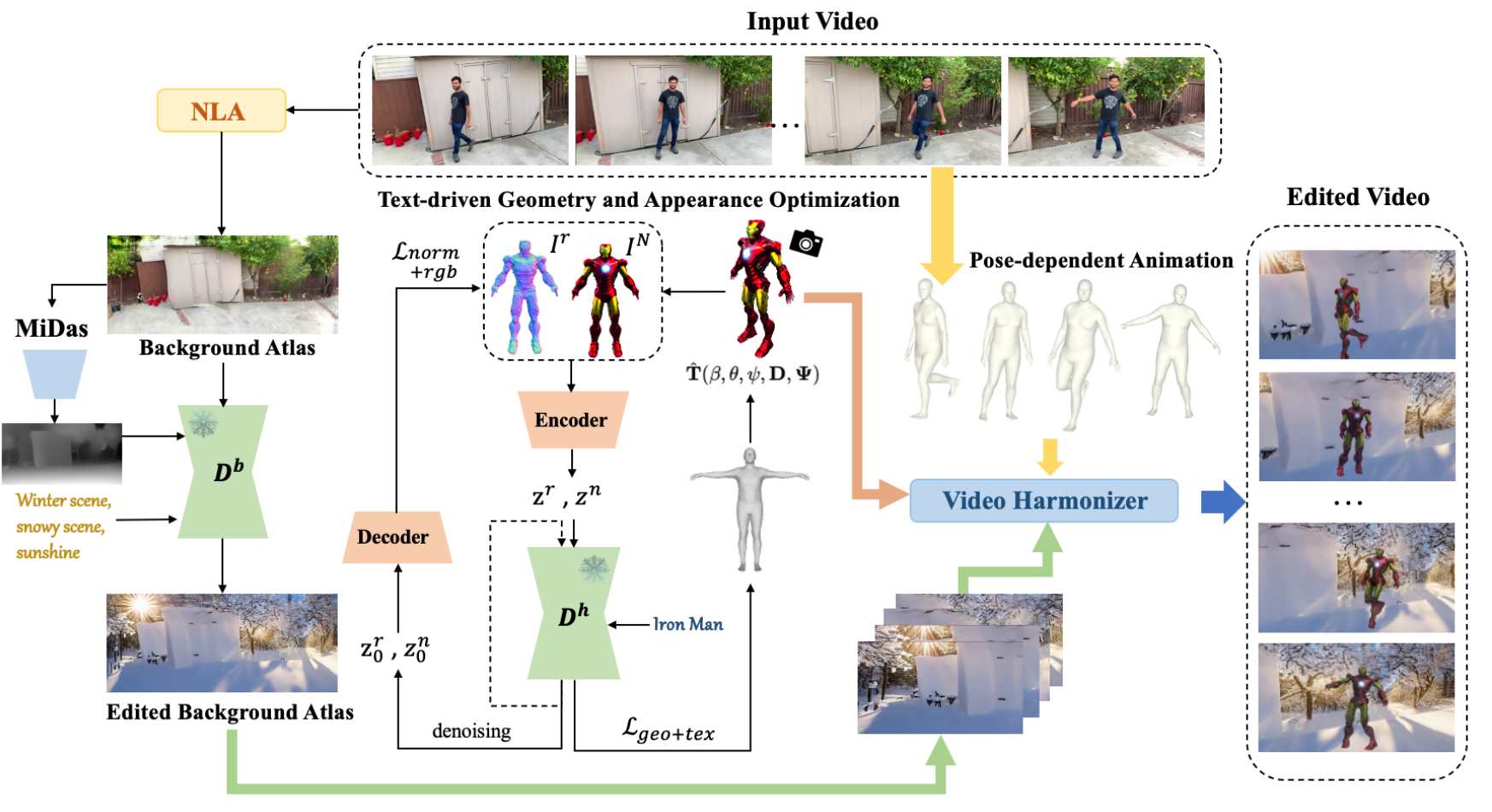}
\caption{\textbf{Framework of DeCo.} The framework is divided into three parts. 1) We utilize NLA \cite{kasten2021layered} to create a background atlas for the input video, with the depth-guided diffusion model $D_b$ generating scenes based on background prompts. 2) The diffusion model $D_h$ optimizes shape parameters $\beta$, expressive parameters $\psi$, displacement layer $\mathbf{D}$, and texture map $\mathbf{\Psi}$ of $\hat{\mathbf{T}}$ using $\mathcal{L}_{\mathrm{geo+tex}}$ and $\mathcal{L}_{\mathrm{norm+rgb}}$. The RGB image $I_r$ and normal map $I^N$ rendered from $\hat{\mathbf{T}}$ are encoded into latent vectors $z^r$ and $z^n$ for $\mathcal{L}_{\mathrm{geo+tex}}$, then denoised to $z^r_0$ and $z^n_0$ for $\mathcal{L}_{\mathrm{norm+rgb}}$. $\hat{\mathbf{T}}$ is animated using pose parameters estimated from the original video. 3) We composite the edited targets with a video harmonizer to ensure the harmonization between them.}
\label{2}
\end{figure}
%


\section{Preliminaries}
\textbf{Latent Diffusion Model (LDM)} 
is an extension of DDPM \cite{ho2020denoising}, which generates images by progressively reducing noise starting from Gaussian noise, aiming to save memory by operating in the latent space. Specifically, LDM, comprising both an encoder and a decoder, encodes an image into a low-resolution latent code $z$, which can subsequently be reconstructed back into the image space with the decoder. The initial latent code $z_0$ evolves into $z_t$ over time step $t$ through the introduction of random noise $\epsilon$, sampled from a standard normal distribution. To reverse this diffusion process, a U-Net \cite{ronneberger2015u} noise predictor $\epsilon_\phi$ parameterized by $\phi$ is trained using the loss function: 
\begin{equation}
    \min _{\phi} \mathbb{E}_{\mathbf{z}_0, \epsilon \sim \mathcal{N}(0, I), t}\left\|\epsilon_{\phi}\left(z_t;y,t\right)-\epsilon\right\|_2^2,
\end{equation}
where $y$ represents a conditional text prompt. 


\textbf{Score Distillation Sampling (SDS)} is utilized to distill knowledge \cite{xiao2024distilling} from a pre-trained diffusion model $\phi$ into a differentiable 3D representation parameterized by $\eta$. Based on LDM, $z$ is encoded from the rendered image $I$, which is generated through $I=\mathbf{g}(\eta)$ based on a given camera pose, with $\mathbf{g}$ representing a differentiable rendering function. SDS loss optimizes $\eta$ by 
computing the gradient of $\mathcal{L}_{\mathrm{SDS}}$ with respect to $z$:
\begin{equation}\label{eq.1}
    \nabla_{\eta} \mathcal{L}_{\mathrm{SDS}}(\phi, z)=\mathbb{E}_{t, {\epsilon}}\left[w(t)\left(\hat{\epsilon}_\phi\left(z_t ; y, t\right)-\epsilon\right) \frac{\partial z}{\partial \eta}\right],
\end{equation}
where $w(t)$ is a time-dependent weighting function that depends on $t$ and $z_t$ is the noised latent vector. Compared to $\epsilon_\phi$, $\hat{\epsilon}_\phi$ incorporates classifier-free guidance \cite{ho2022classifier} to better guide the diffusion process, ensuring alignment with the target prompt. 




\textbf{SMPL-X} is an animatable 3D human model compatible with current rendering engines, which is often represented as canonical deformation and dynamic transformation: 
\begin{equation}\label{eq.2}
\begin{split}
    \mathbf{T}(\beta, \theta, \psi) &= T + B_s(\beta) + B_p(\theta) + B_e(\psi), \\
    \mathbf{M}(\beta, \theta, \psi) &=\mathcal{W}(\mathbf{T}(\beta, \theta, \psi), \mathbf{J}(\beta), \theta, W),
\end{split}
\end{equation}
where \( T \in \mathbb{R}^{10475 \times 3} \) represents the mean shape template in a T-pose. The functions \( B_s \), \( B_p \), and \( B_e \) correspond to shape, pose, and expression blend shapes, respectively. The parameters \( \beta \), \( \theta \), and \( \psi \) define the specific shape, pose, and facial expressions of the human. \( \theta \) encompasses the body pose \( \theta_b \), jaw pose \( \theta_f \), and finger pose \( \theta_h \). $\mathbf{T}(\beta, \theta, \psi)$ can be transformed to the target pose using the linear blend-skinning function $\mathcal{W}$. $\mathbf{J}(\beta)$ takes as inputs $\beta$ to regress 54 joints and $W$ represents skinning weights. 

\section{Methods}

DeCo aims to edit human-centered videos to comply with target text prompts, enabling the generation of dynamic humans with high-fidelity appearances and coherent motions. As illustrated in Fig. \ref{2}, we treat humans and the background as separate editable targets. We represent the background as a layered atlas using NLA and edit it with diffusion models under the guidance of depth map (Sec. \ref{sec.background}). We propose a decoupled dynamic human representation based on a human body prior SMPL-X \cite{loper2023smpl} to generate high-fidelity tailored human meshes through extended SDS loss (Sec. \ref{sec.Decoupled}). Moreover, we introduce a lighting-aware video harmonizer to maintain lighting consistency between edited targets (Sec. \ref{sec.harmonization}).


\subsection{Background Atlas Editing}\label{sec.background}
To propagate the contents with temporal consistency, NLA \cite{kasten2021layered} decomposes a dynamic scene into a collection of atlases, each representing a target for editing. The approach utilizes two coordinate-based MLPs: the first maps pixels in each frame to a UV field, while the second constructs atlases by sampling uv coordinates. In this section, we only consider the construction of the background atlas. Formally, given a video $\textbf{V} = (I_1, I_2, ..., I_N)$ with N denoting the number of frames in the video, we establish UV mapping $\mathcal{M}^{UV}$ and atlas mapping $\mathcal{M}^{A}$ as follows:
\begin{equation}
    \mathrm{uv}_i(\cdot)=\mathcal{M}^{UV}\left(I_i\right), \mathrm{B}_A=\mathcal{M}^{A}(\mathrm{uv}_i(\cdot)),
\end{equation}
where $\mathrm{uv}_i(\cdot)$ represents the uv coordinates for the pixels in the $i$-th frame, and $\mathrm{B}_A$ denotes the background atlas. By mapping $\mathrm{B}_A$ back to the original video for reconstruction loss, we can train these two mapping networks. Once trained, they can be directly applied for editing without further updates. Subsequently, we employ a pre-trained depth-guided diffusion model $D^b$ (e.g., ControlNet \cite{zhang2023adding}) to edit $B_A$, resulting in $\mathrm{B}_A^{\prime}$. The corresponding edited background frames $\textbf{V}^b = (I^b_1, I^b_2, ..., I^b_N)$, maintaining the original video's structure, are obtained by sampling with their respective uv coordinates:

\begin{equation}
 I^b_i=\mathrm{uv}_i(\mathrm{B}_A^{\prime})
\end{equation}

\subsection{Decoupled Dynamic Human Representation}\label{sec.Decoupled}


Non-rigid objects such as humans encounter large deformation during moving, easily leading to distorted appearances and incoherent motions. To tackle this problem, we decouple the dynamic human into text-driven geometry and appearance optimization and pose-dependent animation through explicitly leveraging a human body prior SMPL-X \cite{corona2021smplicit}.   

\textbf{Text-driven Geometry and Appearance Optimization. } With fixed pose parameters $\theta$, we learn the geometry and appearance of the human mesh customized to the text prompt under the canonical pose. To generate expressive human meshes, we add learnable displacements $\mathbf{D}$ on the template of SMPL-X \cite{corona2021smplicit,patel2020tailornet}. Given the sparse distribution of vertices on the SMPL-X body's surface, we refine the mesh by subdividing these vertices and interpolating the skinning weights, similar to \cite{liao2023tada}. Moreover, since the mesh aligned with SMPL-X has fixed topology structure, the texture of it can be defined as a learnable texture map $\mathbf{\Psi}$. Thus, Eq. \ref{eq.2} can be improved as follows:
\begin{equation}
\begin{split}
    \hat{\mathbf{T}}(\beta, \theta, \psi, \mathbf{D},\mathbf{\Psi}) &=\mathcal{S}(\mathbf{T}(\beta, \theta, \psi, \mathbf{\Psi}))+\mathbf{D}, 
\end{split}
\end{equation}
where $\mathcal{S}$ denotes vertices subdivision operation.

We optimize the mesh's learnable parameters using the SDS loss proposed in \cite{poole2022dreamfusion}, which leverages a T2I diffusion model \cite{rombach2022high} to guide the 3D generation, aligning the rendered image with the learned image distribution. However, the SDS loss is calculated in the latent space to save memory costs, which easily tends to converge to inaccurate shapes like cartoon-like bodies without finetuning the T2I diffusion model on realistic human datasets. To address this issue, we propose to extend it in both image and normal space to constrain the generation of geometric shapes and to produce high-fidelity textures during the denoising process. Concretely, with $\hat{\mathbf{T}}$, we first render its RGB image $I^r$ and normal map $I^N$ from a sampled camera pose as the integration of normal maps into the SDS loss has been shown to improve geometric consistency with texture \cite{liao2023tada,huang2023humannorm}. As illustrated in Fig. \ref{2}, $I^r$ and $I^N$ are then encoded into $z^r$ and $z^n$, respectively. Inspired by \cite{zhu2023hifa}, the noise residual of Eq. \ref{eq.1} can be reformulated as latent vector residual. In terms of the noisy normal latent vector $z^n$, we can denoise it using the noise predictor $\hat{\epsilon}_\phi$, and denote the denoised result as $z^n_0$. $z^n_0$ can be formulated as $z^n_0:=\frac{1}{\alpha_t}({z}^n_t-\sigma_t {\hat{\epsilon}_\phi}(z^n_t ; y, t))$ where $\alpha(t)$ and $\sigma_t$ define the noise scheduler. The calculation of geometry loss $\mathcal{L}_{\mathrm{geo}}$ is as follows: 

\begin{equation}
\begin{aligned}
\nabla_\delta \mathcal{L}_{\mathrm{geo}}(\phi, z^n) & = \mathbb{E}_{t, \epsilon}\left[\omega(t)(\hat{\boldsymbol{\epsilon}_\phi}(z^n;y,t)-\epsilon) \frac{\partial z^n}{\partial \delta}\right] \\
& \Rightarrow \mathbb{E}_{t, \epsilon}\left[\omega(t)\left(\frac{1}{\sigma_t}\left(z^n_t-\alpha_t z_0^n\right)-\frac{1}{\sigma_t}\left(z_t^n-\alpha_t z\right)\right) \frac{\partial z^n}{\partial \delta}\right] \\
& = \mathbb{E}_{t, \epsilon}\left[\omega(t) \frac{\alpha_t}{\sigma_t}(z^n-z_0^n) \frac{\partial z^n}{\partial \delta}\right],
\end{aligned}
\end{equation}

where $\delta=\{\beta,\psi,\mathbf{D}\}$. Subsequently, $z^n_0$ is decoded back to normal space to get the reconstructed normal map $\hat{I}^N$. To guide the generation of detailed full-body, we incorporate the normal loss $\mathcal{L}_{\mathrm{norm}}$ into $\mathcal{L}_{\mathrm{geo}}$ by adding the reconstruction loss between $I^N$ and $\hat{I}^N$:

\begin{equation}
\begin{aligned}
\mathcal{L}_{\mathrm{geo+norm}}(\phi, z^n,I^N) & =\mathbb{E}_{t, \epsilon}\left[\omega(t) \frac{\alpha_t}{\sigma_t}(z^n-z_0^n) \frac{\partial z^n}{\partial \delta}\right] + \lambda_n\|I^N-\hat{I}^N\|^2.
\end{aligned}
\end{equation}
Similarly, it can be derived that $z^r_0:=\frac{1}{\alpha_t}({z}^r_t-\sigma_t {\hat{\epsilon}_\phi}(z^r_t ; y, t))$ with respect to the noisy image latent vector $z^r$.
We also reframe the texture loss $\mathcal{L}_{\mathrm{tex}}$ in the form of a latent vector residual:
\begin{equation}
\begin{aligned}
 \nabla_\Psi \mathcal{L}_{\mathrm{tex}}(\phi, z^r) &= \mathbb{E}_{t, \epsilon}\left[\omega(t)(\hat{\boldsymbol{\epsilon}_\phi}(z^r;y,t)-\epsilon) \frac{\partial z^r}{\partial \Psi}\right] \\
 & \Rightarrow \mathbb{E}_{t, \boldsymbol{\epsilon}}\left[\omega(t) \frac{\alpha_t}{\sigma_t}(z^r-z_0^r) \frac{\partial z^r}{\partial \Psi}\right].
\end{aligned}
\end{equation}

After acquiring ${z}_0^r$, we feed it into the decoder to generate the reconstructed image $\hat{I}^r$. To further enhance texture quality, we extend $\mathcal{L}_{\mathrm{tex}}$ to the image space with image loss $\mathcal{L}_{\mathrm{rgb}}$ minimizing the difference between $I^r$ and $\hat{I}^r$:

\begin{equation}
\begin{aligned}
\mathcal{L}_{\mathrm{tex+rgb}}(\phi, z^r,I^r) & =\mathbb{E}_{t, \boldsymbol{\epsilon}}\left[\omega(t) \frac{\alpha_t}{\sigma_t}(z^r-z_0^r) \frac{\partial z^r}{\partial \Psi}\right] + \lambda_r\|I^r-\hat{I}^r\|^2,
\end{aligned}
\end{equation}

where $\lambda_n$ and $\lambda_r$ are the scaling parameters and we set their default value as 0.01. We jointly optimize the geometry and texture of the human mesh $\hat{\mathbf{T}}$ and set $\lambda_{geo}$ and $\lambda_{tex}$ as the corresponding hyper-parameters to represent the total loss $\mathcal{L}_{total}$:

\begin{equation}
    \mathcal{L}_{total} = \lambda_{geo}\mathcal{L}_{geo+norm} + \lambda_{tex}\mathcal{L}_{tex+rgb}.
\end{equation}\label{eq.11}


\textbf{Pose-dependent Animation.} To animate the generated mesh in alignment with the original video's pose, we employ 3D pose estimation approaches to derive pose parameters for the controllable mesh from a monocular video. Specifically, we apply HybrIK \cite{li2021hybrik} to acquire the estimated pose parameters $\hat{\theta}$, which are used for rotating the 3D joints of $\hat{\mathbf{T}}$ under the canonical pose. According to Eq. \ref{eq.2}, the transformed mesh $\hat{\mathbf{M}}$ can be obtained as follows: 

\begin{equation}\label{eq.4}
\begin{split}
    \hat{\mathbf{M}}((\cdot), \hat{\theta}) &=\mathcal{W}(\hat{\mathbf{T}}(\cdot), \mathbf{J}(\hat{\beta}), \hat{\theta}, \hat{W}),
\end{split}
\end{equation}

where $\hat{\beta}$ are shape parameters derived from the geometry generation and are fixed in this stage. $\hat{\mathcal{W}}$ are the interpolated skinning weights corresponding to mesh subdivision. We omit the optimized parameters related to $\hat{\mathbf{T}}$ for the sake of brevity. More details of the pose estimation algorithm used in our paper are provided in the supplementary material. 

\subsection{Lighting-Aware Video Harmonizer}\label{sec.harmonization}

Illumination discrepancies may arise if the edited human is directly rendered into the edited scene. To solve the issue, we propose a simple yet effective lighting-aware video harmonizer to maintain the lighting consistency between the edited targets, which is ignored in the prior methods. We consider the harmonization problem as the combination of the intrinsic consistency of albedo maps and shading, treating an image as the product of material reflectance and scene lighting effects. To harmonize the albedo maps, we employ an off-the-shelf image editing network \cite{miangoleh2023realistic} to regress sets of editing parameters, allowing for adjustments to exposure, white balance and other image attributes. The shading harmonization is divided into two stages. Concretely, we first estimate the shading of the background using the Lambertian shading model, exploiting a lighting model to shade each pixel of the background based on the surface normal, which incorporates a directional light source and a constant ambient illumination. Applying the lighting model to the surface normal of the foreground, we can achieve the initial shading harmonization. Subsequently, we utilize a pre-trained shading refinement network proposed by \cite{careaga2023intrinsic} to refine the global shading, which takes as inputs the foreground mask, the RGB composite image (product of the harmonized albedo map and the initial harmonized shading), the initial shading, the surface normal and the depth map to output the refined shading. Note that the above-mentioned pre-trained models are both trained in the wild datasets, as we can employ them to our task without considering domain gaps. The final harmonized image is the product of the harmonized albedo map and the refined shading. We extend the harmonization to the video by adding an exponential moving average to each frame, which preserves the lighting smoothness across frames.


\section{Experiments}

\subsection{Implementation Details.} We train two mapping networks for background atlas editing through 300,000 iterations of each video. The resolution for the background was set to 768 $\times$ 432, and we employ a DDIM \cite{song2020denoising} sampler with 20 steps. During the stage of text-driven geometry and appearance optimization, we sample the corresponding camera poses separately for rendering the full body and the head. In addition, we utilize hierarchical rendering to render RGB images, scaling from 32 $\times$ 32 to 512 $\times$ 512 while maintaining the rendered normal map at a resolution of 512 $\times$ 512. 
During the joint training of geometry and texture, the geometry is fixed after 50 iterations, while the texture is trained for 150 iterations. The corresponding loss weights, $\lambda_n$ and $\lambda_r$ are both set to 1.

\subsection{Baseline Comparisons.}

\textbf{Dateset.} To evaluate our approach, we employ the videos from Neuman \cite{jiang2022neuman} and Hosnerf \cite{liu2023hosnerf} datasets. The former comprises six videos, each ranging from 10 to 20 seconds in length, while the latter consists of six videos as well, with durations between 60 to 120 seconds each. We also select several human-related videos from 3DPW \cite{von2018recovering} to further evaluate our DeCo. The prompts used for editing are generated from ChatGPT, which are introduced in the supplementary material. 

\textbf{Baselines.} To demonstrate the effectiveness of our proposed DeCo, we perform a comparative analysis against the following baseline methods. 1) \textit{Tune-A-Video} \cite{wu2023tune} is the pioneering work that extends T2I models to video domains. 2) \textit{StableVideo} \cite{chai2023stablevideo} utilizes NLA to decompose the video editing into foreground and background editing, using the object edges and depth maps as the structure guidance. 3) \textit{TokenFlow} \cite{geyer2023tokenflow} maintains the inter-frame correspondences and redundancy observed in the original video for temporal consistency. 4) \textit{Rerender-A-Video} \cite{yang2023rerender} leverages optical flow for cross-frame constraints to enhance the propagation of edited frames. To ensure a fair comparison, for methods requiring tuning a video on the model, such as \textit{Tune-A-Video}, we extract image captions from their key frames using the BLIP model \cite{li2022blip}. Subsequently, we replace phrases related to edited objects and the background with the desired prompt while retaining the rest unchanged.  


\begin{figure}[t]
\centering
\setlength{\abovecaptionskip}{0pt}
\includegraphics[width=1.0\textwidth,height=0.4\textheight]{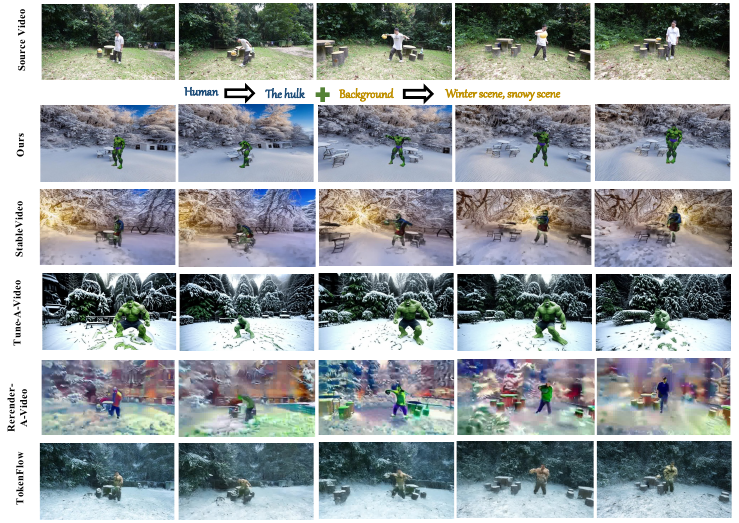}
\caption{\textbf{Qualitative comparisons of our DeCo and four representative video editing methods.} DeCo achieves much more satisfactory results in terms of video quality and motion consistency. }
\label{3}
\end{figure}


\textbf{Qualitative Comparison.} To better demonstrate the motion consistency of the edited video, we showcase five frames that are more than 10 frames apart in Fig. \ref{3}. Tune-A-Video \cite{wu2023tune} can generate coherent frames. However, it struggles to maintain the consistency of human motions from the original video. The results generated by StableVideo \cite{chai2023stablevideo} indicate that utilizing a diffusion model conditioned on canny edges is insufficient for editing complex objects like humans. Additionally, it is unable to modify the shapes of humans, which limits the diversity of the edited human. TokenFlow \cite{geyer2023tokenflow} produces results similar to the original video, with only minor changes in the facial area.
\begin{wrapfigure}[20]{r}{0.5\textwidth} 
  \centering
\includegraphics[width=0.5\textwidth,height=0.26\textheight]{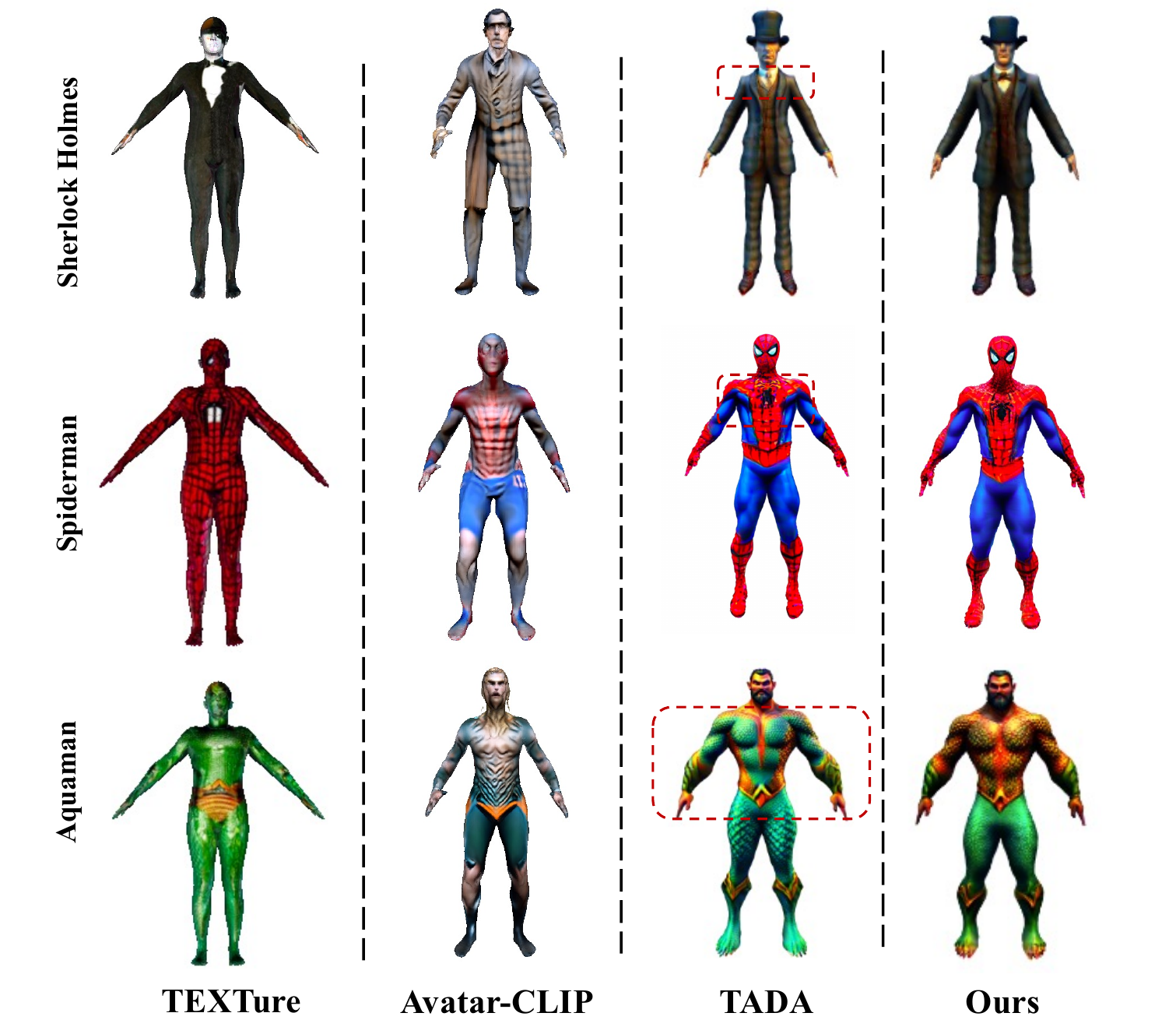}
  \caption{\textbf{Text-guided human generation methods.} TADA \cite{liao2023tada} encounters several issues, such as cartoon-like body shapes ($\textit{Sherlock Holmes}$), blurry patterns ($\textit{Spiderman}$), and misalignment with the prompts ($\textit{Aquaman}$). }
  \label{fig:wrapfig}
\end{wrapfigure}

 The results produced by Rerender-A-Video \cite{yang2023rerender} exhibit blurriness in both the foreground and background. Overall, none of them can simultaneously ensure the quality of the edited video and maintain motion consistency as the source video. In comparison, our DeCo not only generates diverse humans with high-fidelity appearances in response to different prompts but also enables complex animation with target pose sequences, which effectively avoids appearance distortions caused by changes in motion. We further conduct a qualitative analysis comparing our proposed text-guided human generation method with alternative approaches. Note that as we construct the humans based on SMPL-X, our comparison is limited to those human generation methods conditioned on the SMPL model. From Fig. \ref{fig:wrapfig}, TEXTure \cite{richardson2023texture} generates humans with poor texture, failing to accurately depict the human body. AvatarCLIP \cite{hong2022avatarclip} struggles to represent character-specific vertex displacements. While TADA \cite{liao2023tada} is capable of generating diverse humans, it faces challenges in creating humans that are closely aligned with the text, featuring high-fidelity textures and realistic body shapes.



\textbf{Quantitative Comparison.} We quantify DeCo against baselines using CLIP \cite{radford2021learning} metrics across 13 scenes from three datasets, with each scene corresponding to two characters, amounting to a total of 26 videos. To gauge the frame consistency of edited videos, we extract CLIP \cite{radford2021learning} image embeddings for all frames and compute the average cosine similarity among the adjacent frames. To evaluate textual faithfulness, we calculate the CLIP scores by comparing all video frames with their target prompts. As shown in Tab. \ref{compare}, our method achieves the best frame consistency and textual faithfulness. We also conduct a user study to evaluate DeCo. We select 13 groups of videos, each corresponding to a different scene. For each group, users are shown the original video alongside its corresponding target prompts, followed by five edited videos that are generated by various methods and randomly arranged. We invite 15 volunteers to rank each group of videos based on their temporal consistency, editing ability, and image quality. The scoring system ranges from 5, being the highest, to 1, the lowest, with no repeated scores allowed. The final report will present the average scores across all groups. From Tab. \ref{compare}, our method achieves the highest human preference in all aspects.

\begin{table*}[t]
\tiny
    \centering
    \resizebox{0.95\linewidth}{!}{
    \begin{tabular}{c c|c c c c c}
    \toprule[0.5pt]
    \multicolumn{2}{c|}{Metrics} & TAV & StableVideo & TokenFlow & RAV  & DeCo \\
    \bottomrule[0.pt]
    \toprule[0.5pt]
    \multirow{2}{*}{CLIP $\uparrow$} & Fram-Con & 93.58 & 93.55& 91.13 & 96.76 & \textbf{96.91} \\
	& Tex-Faith & 32.35 & 29.40 & 29.99 & 32.22 & \textbf{35.25} \\ 
    \midrule[0.5pt]
    \multirow{3}{*}{User $\uparrow$} & Temporal & 2.23 & 3.23 & 2.54 & 3.08 & \textbf{3.92} \\
	& Edit & 3.23  & 2.54 &2.46  & 2.62 & \textbf{4.15} \\ 
        & Quality & 2.62 & 2.85 & 2.77 & 2.54 & \textbf{4.23} \\
   \bottomrule[0.5pt]
    \end{tabular}
	}
	
    \caption{\label{compare} Quantitative evaluation of the results obtained from different approaches, including TAV \cite{wu2023tune}, StableVideo \cite{chai2023stablevideo}, TokenFlow \cite{geyer2023tokenflow}, RAV \cite{yang2023rerender}, and DeCo. $`Frame-Con`$ and $`Tex-Faith`$ respectively denote frame consistency and texture faithfulness. $`Temporal`$,$`Edit`$ and $`Quality`$ correspond to user evaluations of temporal consistency, editing capability, and image quality, respectively. The best scores are highlighted in \textbf{bold}.}
\end{table*}

 
\begin{figure}[t] 
\centering
\setlength{\abovecaptionskip}{0pt}
\includegraphics[width=1.0\textwidth,height=0.24\textheight]{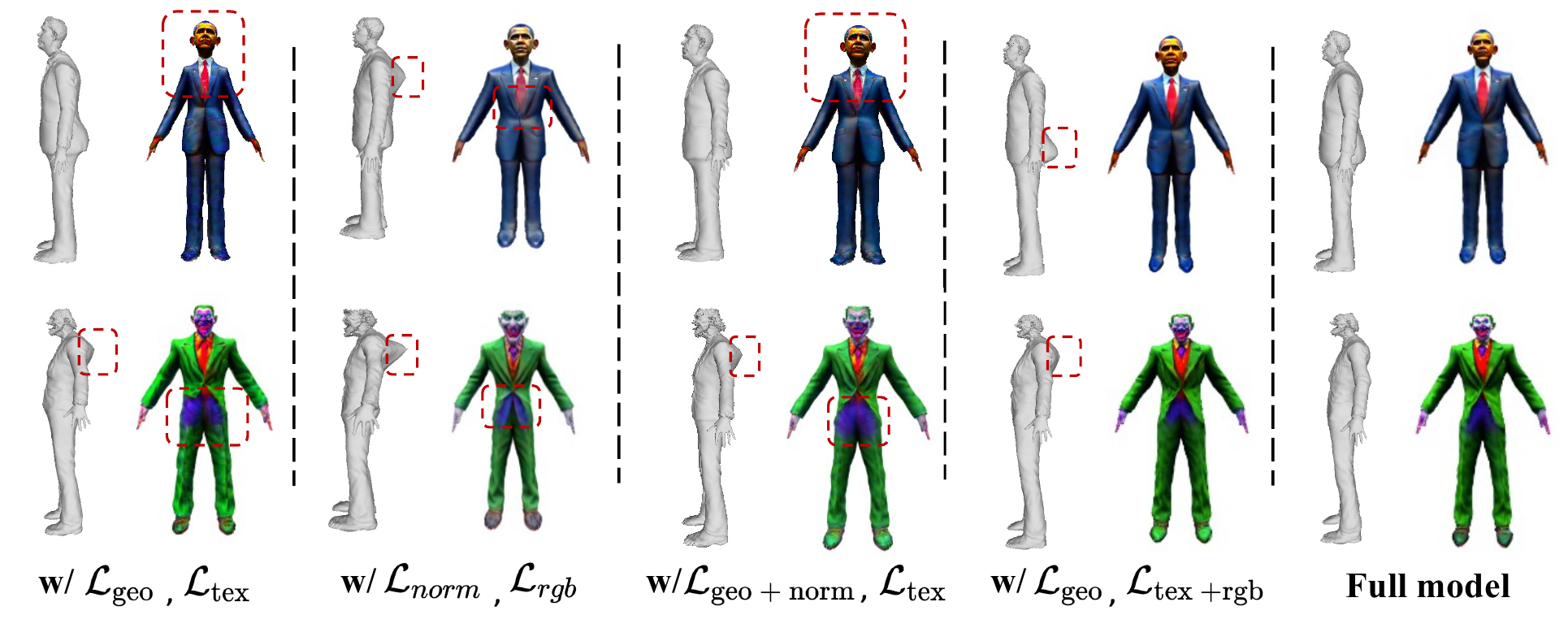}
\caption{\textbf{Effects of the loss functions.} The two prompts used from top to bottom are "Barack Obama" and "Joker". Unnatural head-to-shoulder ratio, blurred textures, and distorted body shapes are shown in red boxes.}
\label{5}
\end{figure} 

\begin{figure}[t]
\centering
\setlength{\abovecaptionskip}{0pt}
\includegraphics[width=1.0\textwidth,height=0.24\textheight]{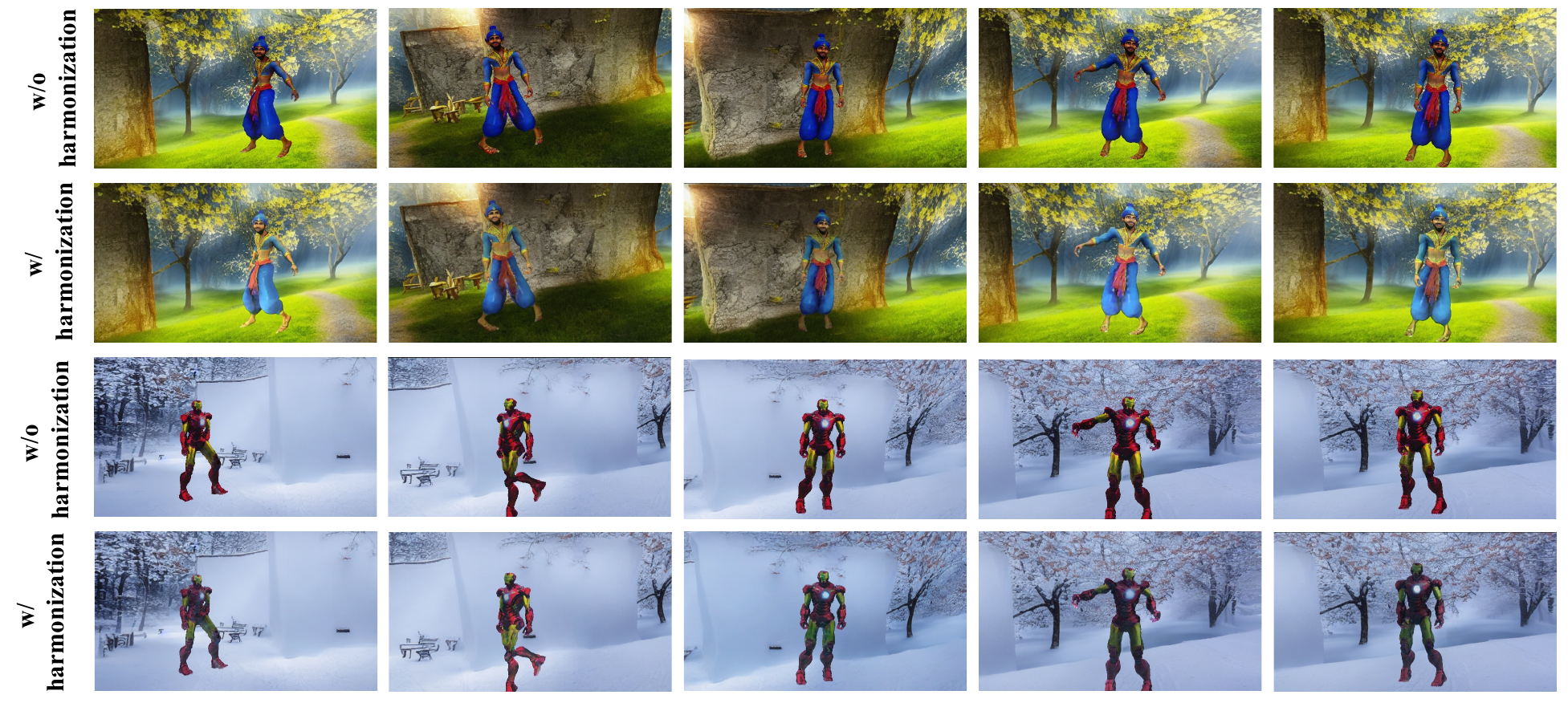}
\caption{\textbf{Effects of the lighting-aware video harmonizer.} The prompts used, from top to bottom, are "Aladdin" with
"spring scene, forest scene" and "Iron man" with "winter scene, snowy scene". }
\label{6}
\end{figure}


\subsection{Ablation Study}

\textbf{Extending SDS loss to normal and image space.} As illustrated in Fig. \ref{5}, the \textit{Obama} in the first column exhibits an unrealistic head-to-shoulder ratio, making it appear less authentic. Moreover, the patterns of the Jokers are blurry. Additionally, as observed in the second column, employing only the reconstruction loss in the normal and image spaces yields suboptimal results. We argue that this is due to performing only a single step in the denoising process, which does not ensure high-quality denoised noises. However, the guidance provided by the SDS loss facilitates a more rapid convergence to a plausible human shape and appearance. The third column demonstrates that incorporating $\mathcal{L}_{norm}$ into geometry learning results in a realistic body shape. As shown in the fourth column, integrating $\mathcal{L}_{rgb}$ into the texture learning generates sharper textures. In the fifth column, we achieve lifelike humans with high-fidelity textures. 

\textbf{Video Harmonizer.} As illustrated in Fig. \ref{6}, the first row demonstrates a noticeable inconsistency between \textit{Aladdin's} brightness and the background when the video harmonizer is not applied. Note that the second and third columns reveal that our proposed video harmonizer is aware of lighting occlusions, effectively portraying shadows when sunlight is obstructed. In the second video, the harmonizer adjusts \textit{Iron man} to a cool tone, aligning with the lighting expected in a snowy scene. With the incorporation of the video harmonizer, the frame achieves lighting consistency.

\section{Limitations and Future Works}
For each video, our method is required to train its atlas mapping networks individually, which introduces additional time costs. In the future, we aim to discover more efficient zero-shot methods for representing video foreground and background. Furthermore, our human pose parameters are derived from existing 3D pose estimation techniques, inheriting their limitations. 

\section{Conclusion}
We propose an innovative video editing framework designed for independent editing of humans and backgrounds to maintain spatial-temporal consistency. By leveraging a parametric human prior, our approach decouples dynamic humans from motions, enabling the generation of diverse shapes and appearances under the canonical pose. We extend the SDS loss to the image and normal space to enhance the geometry and texture details. Additionally, a novel lighting-aware video harmonizer addresses the inconsistent lighting problem between edited targets.

\section*{Acknowledgments}
This work is supported by National Natural Science Foundation of China (NSFC) 62272172, Guangdong Basic and Applied Basic Research Foundation 2023A1515012920, Zhuhai Science and Technology Plan Project(2320004002758). This research is also supported by the MoE AcRF Tier 1 grant (RG14/22).

%
%
\bibliographystyle{splncs04}
\bibliography{main}

\begin{thebibliography}{10}
\providecommand{\url}[1]{\texttt{#1}}
\providecommand{\urlprefix}{URL }
\providecommand{\doi}[1]{https://doi.org/#1}

\bibitem{bar2022text2live}
Bar-Tal, O., Ofri-Amar, D., Fridman, R., Kasten, Y., Dekel, T.: Text2live: Text-driven layered image and video editing. In: European conference on computer vision. pp. 707--723. Springer (2022)

\bibitem{brooks2023instructpix2pix}
Brooks, T., Holynski, A., Efros, A.A.: Instructpix2pix: Learning to follow image editing instructions. In: Proceedings of the IEEE/CVF Conference on Computer Vision and Pattern Recognition. pp. 18392--18402 (2023)

\bibitem{cai2023smpler}
Cai, Z., Yin, W., Zeng, A., Wei, C., Sun, Q., Wang, Y., Pang, H.E., Mei, H., Zhang, M., Zhang, L., et~al.: Smpler-x: Scaling up expressive human pose and shape estimation. arXiv preprint arXiv:2309.17448  (2023)

\bibitem{cao2023dreamavatar}
Cao, Y., Cao, Y.P., Han, K., Shan, Y., Wong, K.Y.K.: Dreamavatar: Text-and-shape guided 3d human avatar generation via diffusion models. arXiv preprint arXiv:2304.00916  (2023)

\bibitem{careaga2023intrinsic}
Careaga, C., Miangoleh, S.M.H., Aksoy, Y.: Intrinsic harmonization for illumination-aware compositing. arXiv preprint arXiv:2312.03698  (2023)

\bibitem{ceylan2023pix2video}
Ceylan, D., Huang, C.H.P., Mitra, N.J.: Pix2video: Video editing using image diffusion. In: Proceedings of the IEEE/CVF International Conference on Computer Vision. pp. 23206--23217 (2023)

\bibitem{chai2023stablevideo}
Chai, W., Guo, X., Wang, G., Lu, Y.: Stablevideo: Text-driven consistency-aware diffusion video editing. In: Proceedings of the IEEE/CVF International Conference on Computer Vision. pp. 23040--23050 (2023)

\bibitem{chen2023fantasia3d}
Chen, R., Chen, Y., Jiao, N., Jia, K.: Fantasia3d: Disentangling geometry and appearance for high-quality text-to-3d content creation. arXiv preprint arXiv:2303.13873  (2023)

\bibitem{chen2023control}
Chen, W., Wu, J., Xie, P., Wu, H., Li, J., Xia, X., Xiao, X., Lin, L.: Control-a-video: Controllable text-to-video generation with diffusion models. arXiv preprint arXiv:2305.13840  (2023)

\bibitem{chen2023it3d}
Chen, Y., Zhang, C., Yang, X., Cai, Z., Yu, G., Yang, L., Lin, G.: It3d: Improved text-to-3d generation with explicit view synthesis. arXiv preprint arXiv:2308.11473  (2023)

\bibitem{chu2023video}
Chu, E., Lin, S.Y., Chen, J.C.: Video controlnet: Towards temporally consistent synthetic-to-real video translation using conditional image diffusion models. arXiv preprint arXiv:2305.19193  (2023)

\bibitem{corona2021smplicit}
Corona, E., Pumarola, A., Alenya, G., Pons-Moll, G., Moreno-Noguer, F.: Smplicit: Topology-aware generative model for clothed people. In: Proceedings of the IEEE/CVF conference on computer vision and pattern recognition. pp. 11875--11885 (2021)

\bibitem{esser2023structure}
Esser, P., Chiu, J., Atighehchian, P., Granskog, J., Germanidis, A.: Structure and content-guided video synthesis with diffusion models. In: Proceedings of the IEEE/CVF International Conference on Computer Vision. pp. 7346--7356 (2023)

\bibitem{geyer2023tokenflow}
Geyer, M., Bar-Tal, O., Bagon, S., Dekel, T.: Tokenflow: Consistent diffusion features for consistent video editing. arXiv preprint arXiv:2307.10373  (2023)

\bibitem{gu2023videoswap}
Gu, Y., Zhou, Y., Wu, B., Yu, L., Liu, J.W., Zhao, R., Wu, J.Z., Zhang, D.J., Shou, M.Z., Tang, K.: Videoswap: Customized video subject swapping with interactive semantic point correspondence. arXiv preprint arXiv:2312.02087  (2023)

\bibitem{guo2021image}
Guo, Z., Guo, D., Zheng, H., Gu, Z., Zheng, B., Dong, J.: Image harmonization with transformer. In: Proceedings of the IEEE/CVF international conference on computer vision. pp. 14870--14879 (2021)

\bibitem{guo2021intrinsic}
Guo, Z., Zheng, H., Jiang, Y., Gu, Z., Zheng, B.: Intrinsic image harmonization. In: Proceedings of the IEEE/CVF Conference on Computer Vision and Pattern Recognition. pp. 16367--16376 (2021)

\bibitem{hang2022scs}
Hang, Y., Xia, B., Yang, W., Liao, Q.: Scs-co: Self-consistent style contrastive learning for image harmonization. In: Proceedings of the IEEE/CVF Conference on Computer Vision and Pattern Recognition. pp. 19710--19719 (2022)

\bibitem{hertz2022prompt}
Hertz, A., Mokady, R., Tenenbaum, J., Aberman, K., Pritch, Y., Cohen-Or, D.: Prompt-to-prompt image editing with cross attention control. arXiv preprint arXiv:2208.01626  (2022)

\bibitem{ho2020denoising}
Ho, J., Jain, A., Abbeel, P.: Denoising diffusion probabilistic models. Advances in neural information processing systems  \textbf{33},  6840--6851 (2020)

\bibitem{ho2022classifier}
Ho, J., Salimans, T.: Classifier-free diffusion guidance. arXiv preprint arXiv:2207.12598  (2022)

\bibitem{hong2022avatarclip}
Hong, F., Zhang, M., Pan, L., Cai, Z., Yang, L., Liu, Z.: Avatarclip: Zero-shot text-driven generation and animation of 3d avatars. arXiv preprint arXiv:2205.08535  (2022)

\bibitem{huang2019temporally}
Huang, H.Z., Xu, S.Z., Cai, J.X., Liu, W., Hu, S.M.: Temporally coherent video harmonization using adversarial networks. IEEE Transactions on Image Processing  \textbf{29},  214--224 (2019)

\bibitem{huang2023humannorm}
Huang, X., Shao, R., Zhang, Q., Zhang, H., Feng, Y., Liu, Y., Wang, Q.: Humannorm: Learning normal diffusion model for high-quality and realistic 3d human generation. arXiv preprint arXiv:2310.01406  (2023)

\bibitem{huang2023dreamwaltz}
Huang, Y., Wang, J., Zeng, A., Cao, H., Qi, X., Shi, Y., Zha, Z.J., Zhang, L.: Dreamwaltz: Make a scene with complex 3d animatable avatars. arXiv preprint arXiv:2305.12529  (2023)

\bibitem{jeong2023ground}
Jeong, H., Ye, J.C.: Ground-a-video: Zero-shot grounded video editing using text-to-image diffusion models. arXiv preprint arXiv:2310.01107  (2023)

\bibitem{jiang2023avatarcraft}
Jiang, R., Wang, C., Zhang, J., Chai, M., He, M., Chen, D., Liao, J.: Avatarcraft: Transforming text into neural human avatars with parameterized shape and pose control. arXiv preprint arXiv:2303.17606  (2023)

\bibitem{jiang2022neuman}
Jiang, W., Yi, K.M., Samei, G., Tuzel, O., Ranjan, A.: Neuman: Neural human radiance field from a single video. In: European Conference on Computer Vision. pp. 402--418. Springer (2022)

\bibitem{jiang2021ssh}
Jiang, Y., Zhang, H., Zhang, J., Wang, Y., Lin, Z., Sunkavalli, K., Chen, S., Amirghodsi, S., Kong, S., Wang, Z.: Ssh: A self-supervised framework for image harmonization. In: Proceedings of the IEEE/CVF International Conference on Computer Vision. pp. 4832--4841 (2021)

\bibitem{kara2023rave}
Kara, O., Kurtkaya, B., Yesiltepe, H., Rehg, J.M., Yanardag, P.: Rave: Randomized noise shuffling for fast and consistent video editing with diffusion models. arXiv preprint arXiv:2312.04524  (2023)

\bibitem{kasten2021layered}
Kasten, Y., Ofri, D., Wang, O., Dekel, T.: Layered neural atlases for consistent video editing. ACM Transactions on Graphics (TOG)  \textbf{40}(6),  1--12 (2021)

\bibitem{kawar2023imagic}
Kawar, B., Zada, S., Lang, O., Tov, O., Chang, H., Dekel, T., Mosseri, I., Irani, M.: Imagic: Text-based real image editing with diffusion models. In: Proceedings of the IEEE/CVF Conference on Computer Vision and Pattern Recognition. pp. 6007--6017 (2023)

\bibitem{lee2023shape}
Lee, Y.C., Jang, J.Z.G., Chen, Y.T., Qiu, E., Huang, J.B.: Shape-aware text-driven layered video editing. In: Proceedings of the IEEE/CVF Conference on Computer Vision and Pattern Recognition. pp. 14317--14326 (2023)

\bibitem{li2021hybrik}
Li, J., Xu, C., Chen, Z., Bian, S., Yang, L., Lu, C.: Hybrik: A hybrid analytical-neural inverse kinematics solution for 3d human pose and shape estimation. In: Proceedings of the IEEE/CVF conference on computer vision and pattern recognition. pp. 3383--3393 (2021)

\bibitem{li2022blip}
Li, J., Li, D., Xiong, C., Hoi, S.: Blip: Bootstrapping language-image pre-training for unified vision-language understanding and generation. In: International Conference on Machine Learning. pp. 12888--12900. PMLR (2022)

\bibitem{li2023sweetdreamer}
Li, W., Chen, R., Chen, X., Tan, P.: Sweetdreamer: Aligning geometric priors in 2d diffusion for consistent text-to-3d. arXiv preprint arXiv:2310.02596  (2023)

\bibitem{liang2023flowvid}
Liang, F., Wu, B., Wang, J., Yu, L., Li, K., Zhao, Y., Misra, I., Huang, J.B., Zhang, P., Vajda, P., et~al.: Flowvid: Taming imperfect optical flows for consistent video-to-video synthesis. arXiv preprint arXiv:2312.17681  (2023)

\bibitem{liao2023tada}
Liao, T., Yi, H., Xiu, Y., Tang, J., Huang, Y., Thies, J., Black, M.J.: Tada! text to animatable digital avatars. arXiv preprint arXiv:2308.10899  (2023)

\bibitem{liu2023hosnerf}
Liu, J.W., Cao, Y.P., Yang, T., Xu, Z., Keppo, J., Shan, Y., Qie, X., Shou, M.Z.: Hosnerf: Dynamic human-object-scene neural radiance fields from a single video. In: Proceedings of the IEEE/CVF International Conference on Computer Vision. pp. 18483--18494 (2023)

\bibitem{liu2023video}
Liu, S., Zhang, Y., Li, W., Lin, Z., Jia, J.: Video-p2p: Video editing with cross-attention control. arXiv preprint arXiv:2303.04761  (2023)

\bibitem{loper2023smpl}
Loper, M., Mahmood, N., Romero, J., Pons-Moll, G., Black, M.J.: Smpl: A skinned multi-person linear model. In: Seminal Graphics Papers: Pushing the Boundaries, Volume 2, pp. 851--866 (2023)

\bibitem{lu2022deep}
Lu, X., Huang, S., Niu, L., Cong, W., Zhang, L.: Deep video harmonization with color mapping consistency. arXiv preprint arXiv:2205.00687  (2022)

\bibitem{ma2023x}
Ma, Y., Zhang, X., Sun, X., Ji, J., Wang, H., Jiang, G., Zhuang, W., Ji, R.: X-mesh: Towards fast and accurate text-driven 3d stylization via dynamic textual guidance. In: Proceedings of the IEEE/CVF International Conference on Computer Vision. pp. 2749--2760 (2023)

\bibitem{ma2023follow}
Ma, Y., He, Y., Cun, X., Wang, X., Shan, Y., Li, X., Chen, Q.: Follow your pose: Pose-guided text-to-video generation using pose-free videos. arXiv preprint arXiv:2304.01186  (2023)

\bibitem{metzer2023latent}
Metzer, G., Richardson, E., Patashnik, O., Giryes, R., Cohen-Or, D.: Latent-nerf for shape-guided generation of 3d shapes and textures. In: Proceedings of the IEEE/CVF Conference on Computer Vision and Pattern Recognition. pp. 12663--12673 (2023)

\bibitem{miangoleh2023realistic}
Miangoleh, S.M.H., Bylinskii, Z., Kee, E., Shechtman, E., Aksoy, Y.: Realistic saliency guided image enhancement. In: Proceedings of the IEEE/CVF Conference on Computer Vision and Pattern Recognition. pp. 186--194 (2023)

\bibitem{michel2022text2mesh}
Michel, O., Bar-On, R., Liu, R., Benaim, S., Hanocka, R.: Text2mesh: Text-driven neural stylization for meshes. In: Proceedings of the IEEE/CVF Conference on Computer Vision and Pattern Recognition. pp. 13492--13502 (2022)

\bibitem{mildenhall2021nerf}
Mildenhall, B., Srinivasan, P.P., Tancik, M., Barron, J.T., Ramamoorthi, R., Ng, R.: Nerf: Representing scenes as neural radiance fields for view synthesis. Communications of the ACM  \textbf{65}(1),  99--106 (2021)

\bibitem{mokady2023null}
Mokady, R., Hertz, A., Aberman, K., Pritch, Y., Cohen-Or, D.: Null-text inversion for editing real images using guided diffusion models. In: Proceedings of the IEEE/CVF Conference on Computer Vision and Pattern Recognition. pp. 6038--6047 (2023)

\bibitem{ouyang2023codef}
Ouyang, H., Wang, Q., Xiao, Y., Bai, Q., Zhang, J., Zheng, K., Zhou, X., Chen, Q., Shen, Y.: Codef: Content deformation fields for temporally consistent video processing. arXiv preprint arXiv:2308.07926  (2023)

\bibitem{parmar2023zero}
Parmar, G., Kumar~Singh, K., Zhang, R., Li, Y., Lu, J., Zhu, J.Y.: Zero-shot image-to-image translation. In: ACM SIGGRAPH 2023 Conference Proceedings. pp. 1--11 (2023)

\bibitem{patel2020tailornet}
Patel, C., Liao, Z., Pons-Moll, G.: Tailornet: Predicting clothing in 3d as a function of human pose, shape and garment style. In: Proceedings of the IEEE/CVF conference on computer vision and pattern recognition. pp. 7365--7375 (2020)

\bibitem{pavlakos2019expressive}
Pavlakos, G., Choutas, V., Ghorbani, N., Bolkart, T., Osman, A.A., Tzionas, D., Black, M.J.: Expressive body capture: 3d hands, face, and body from a single image. In: Proceedings of the IEEE/CVF conference on computer vision and pattern recognition. pp. 10975--10985 (2019)

\bibitem{pont20172017}
Pont-Tuset, J., Perazzi, F., Caelles, S., Arbel{\'a}ez, P., Sorkine-Hornung, A., Van~Gool, L.: The 2017 davis challenge on video object segmentation. arXiv preprint arXiv:1704.00675  (2017)

\bibitem{poole2022dreamfusion}
Poole, B., Jain, A., Barron, J.T., Mildenhall, B.: Dreamfusion: Text-to-3d using 2d diffusion. arXiv preprint arXiv:2209.14988  (2022)

\bibitem{qi2023fatezero}
Qi, C., Cun, X., Zhang, Y., Lei, C., Wang, X., Shan, Y., Chen, Q.: Fatezero: Fusing attentions for zero-shot text-based video editing. arXiv preprint arXiv:2303.09535  (2023)

\bibitem{qiu2023psvt}
Qiu, Z., Yang, Q., Wang, J., Feng, H., Han, J., Ding, E., Xu, C., Fu, D., Wang, J.: Psvt: End-to-end multi-person 3d pose and shape estimation with progressive video transformers. In: Proceedings of the IEEE/CVF Conference on Computer Vision and Pattern Recognition. pp. 21254--21263 (2023)

\bibitem{radford2021learning}
Radford, A., Kim, J.W., Hallacy, C., Ramesh, A., Goh, G., Agarwal, S., Sastry, G., Askell, A., Mishkin, P., Clark, J., et~al.: Learning transferable visual models from natural language supervision. In: International conference on machine learning. pp. 8748--8763. PMLR (2021)

\bibitem{ranftl2020towards}
Ranftl, R., Lasinger, K., Hafner, D., Schindler, K., Koltun, V.: Towards robust monocular depth estimation: Mixing datasets for zero-shot cross-dataset transfer. IEEE transactions on pattern analysis and machine intelligence  \textbf{44}(3),  1623--1637 (2020)

\bibitem{richardson2023texture}
Richardson, E., Metzer, G., Alaluf, Y., Giryes, R., Cohen-Or, D.: Texture: Text-guided texturing of 3d shapes. arXiv preprint arXiv:2302.01721  (2023)

\bibitem{rombach2022high}
Rombach, R., Blattmann, A., Lorenz, D., Esser, P., Ommer, B.: High-resolution image synthesis with latent diffusion models. In: Proceedings of the IEEE/CVF conference on computer vision and pattern recognition. pp. 10684--10695 (2022)

\bibitem{ronneberger2015u}
Ronneberger, O., Fischer, P., Brox, T.: U-net: Convolutional networks for biomedical image segmentation. In: Medical Image Computing and Computer-Assisted Intervention--MICCAI 2015: 18th International Conference, Munich, Germany, October 5-9, 2015, Proceedings, Part III 18. pp. 234--241. Springer (2015)

\bibitem{shen2021deep}
Shen, T., Gao, J., Yin, K., Liu, M.Y., Fidler, S.: Deep marching tetrahedra: a hybrid representation for high-resolution 3d shape synthesis. Advances in Neural Information Processing Systems  \textbf{34},  6087--6101 (2021)

\bibitem{sofiiuk2021foreground}
Sofiiuk, K., Popenova, P., Konushin, A.: Foreground-aware semantic representations for image harmonization. In: Proceedings of the IEEE/CVF Winter Conference on Applications of Computer Vision. pp. 1620--1629 (2021)

\bibitem{sohl2015deep}
Sohl-Dickstein, J., Weiss, E., Maheswaranathan, N., Ganguli, S.: Deep unsupervised learning using nonequilibrium thermodynamics. In: International conference on machine learning. pp. 2256--2265. PMLR (2015)

\bibitem{song2020denoising}
Song, J., Meng, C., Ermon, S.: Denoising diffusion implicit models. arXiv preprint arXiv:2010.02502  (2020)

\bibitem{song2019generative}
Song, Y., Ermon, S.: Generative modeling by estimating gradients of the data distribution. Advances in neural information processing systems  \textbf{32} (2019)

\bibitem{tan2023deep}
Tan, L., Li, J., Niu, L., Zhang, L.: Deep image harmonization in dual color spaces. In: Proceedings of the 31st ACM International Conference on Multimedia. pp. 2159--2167 (2023)

\bibitem{tumanyan2023plug}
Tumanyan, N., Geyer, M., Bagon, S., Dekel, T.: Plug-and-play diffusion features for text-driven image-to-image translation. In: Proceedings of the IEEE/CVF Conference on Computer Vision and Pattern Recognition. pp. 1921--1930 (2023)

\bibitem{von2018recovering}
Von~Marcard, T., Henschel, R., Black, M.J., Rosenhahn, B., Pons-Moll, G.: Recovering accurate 3d human pose in the wild using imus and a moving camera. In: Proceedings of the European conference on computer vision (ECCV). pp. 601--617 (2018)

\bibitem{wang2023semi}
Wang, K., Gharbi, M., Zhang, H., Xia, Z., Shechtman, E.: Semi-supervised parametric real-world image harmonization. In: Proceedings of the IEEE/CVF Conference on Computer Vision and Pattern Recognition. pp. 5927--5936 (2023)

\bibitem{wang2023videocomposer}
Wang, X., Yuan, H., Zhang, S., Chen, D., Wang, J., Zhang, Y., Shen, Y., Zhao, D., Zhou, J.: Videocomposer: Compositional video synthesis with motion controllability. arXiv preprint arXiv:2306.02018  (2023)

\bibitem{wei2022capturing}
Wei, W.L., Lin, J.C., Liu, T.L., Liao, H.Y.M.: Capturing humans in motion: Temporal-attentive 3d human pose and shape estimation from monocular video. In: Proceedings of the IEEE/CVF Conference on Computer Vision and Pattern Recognition. pp. 13211--13220 (2022)

\bibitem{wu2023tune}
Wu, J.Z., Ge, Y., Wang, X., Lei, S.W., Gu, Y., Shi, Y., Hsu, W., Shan, Y., Qie, X., Shou, M.Z.: Tune-a-video: One-shot tuning of image diffusion models for text-to-video generation. In: Proceedings of the IEEE/CVF International Conference on Computer Vision. pp. 7623--7633 (2023)

\bibitem{xiao2024distilling}
Xiao, Y., Wang, D., Li, B., Wang, M., Wu, X., Zhou, C., Zhou, Y.: Distilling autoregressive models to obtain high-performance non-autoregressive solvers for vehicle routing problems with faster inference speed (2024)

\bibitem{yang2023rerender}
Yang, S., Zhou, Y., Liu, Z., Loy, C.C.: Rerender a video: Zero-shot text-guided video-to-video translation. arXiv preprint arXiv:2306.07954  (2023)

\bibitem{yang2023lods}
Yang, X., Chen, Y., Chen, C., Zhang, C., Xu, Y., Yang, X., Liu, F., Lin, G.: Learn to optimize denoising scores: A unified and improved diffusion prior for 3d generation. arXiv:2312.04820  (2023)

\bibitem{Yang_Liu_Xu_Su_Wu_Lin_2024}
Yang, X., Liu, F., Xu, Y., Su, H., Wu, Q., Lin, G.: Diverse and stable 2d diffusion guided text to 3d generation with noise recalibration. Proceedings of the AAAI Conference on Artificial Intelligence  \textbf{38}(7),  6549--6557 (Mar 2024). \doi{10.1609/aaai.v38i7.28476}, \url{https://ojs.aaai.org/index.php/AAAI/article/view/28476}

\bibitem{yu2023gla}
Yu, B.X., Zhang, Z., Liu, Y., Zhong, S.h., Liu, Y., Chen, C.W.: Gla-gcn: global-local adaptive graph convolutional network for 3d human pose estimation from monocular video. In: Proceedings of the IEEE/CVF International Conference on Computer Vision. pp. 8818--8829 (2023)

\bibitem{zhang2023avatarverse}
Zhang, H., Chen, B., Yang, H., Qu, L., Wang, X., Chen, L., Long, C., Zhu, F., Du, K., Zheng, M.: Avatarverse: High-quality \& stable 3d avatar creation from text and pose. arXiv preprint arXiv:2308.03610  (2023)

\bibitem{zhang2023adding}
Zhang, L., Rao, A., Agrawala, M.: Adding conditional control to text-to-image diffusion models. In: Proceedings of the IEEE/CVF International Conference on Computer Vision. pp. 3836--3847 (2023)

\bibitem{zhang2023controlvideo}
Zhang, Y., Wei, Y., Jiang, D., Zhang, X., Zuo, W., Tian, Q.: Controlvideo: Training-free controllable text-to-video generation. arXiv preprint arXiv:2305.13077  (2023)

\bibitem{zhong2021mv}
Zhong, X., Wu, Z., Tan, T., Lin, G., Wu, Q.: Mv-ton: Memory-based video virtual try-on network. In: Proceedings of the 29th ACM International Conference on Multimedia. pp. 908--916 (2021)

\bibitem{zhu2023hifa}
Zhu, J., Zhuang, P.: Hifa: High-fidelity text-to-3d with advanced diffusion guidance. arXiv preprint arXiv:2305.18766  (2023)

\end{thebibliography}
\end{document}


\title{Supplementary Materials for DeCo: Decoupled Human-Centered Diffusion Video Editing with Motion Consistency} 

\author{Xiaojing Zhong\inst{1,2} \and
Xinyi Huang\inst{1} \and
Xiaofeng Yang\inst{2} \and Guosheng Lin\inst{2}\thanks{Corresponding Authors} \and Qingyao Wu\inst{1,3,\footnotemark[1]}}

\authorrunning{X.Zhong et al.}

\institute{School of Software Engineering, South China University of Technology \\
 \and 
Nanyang Technological University\\ 
\and 
Peng Cheng Laboratory \\
\email{vzxj12@gmail.com, gslin@ntu.edu.sg, qyw@scut.edu.cn}}

\titlerunning{Supp-DeCo}




\maketitle
\section{Pose-dependent animation}

In this stage, the goal is to estimate the pose parameters $\hat{\theta}$ from the original video, while keeping parameters optimized in geometry and texture optimization fixed. Following HybrIK \cite{li2021hybrik}, we employ a ResNet network to extract features from each video frame. These features are then processed through deconvolution layers to generate 3D heatmaps, which are subsequently converted into 3D poses $\textbf{P}$ using a softargmax function. Concurrently, the features are fed into MLP layers to estimate twist angles $\textbf{S}$. The canonical human template $\hat{\textbf{T}}$, which characterizes the desired body shape and appearance under the rest pose, is obtained from the geometry and texture optimization stage. Finally, HybrIK takes $\textbf{P}$, $\textbf{S}$, and $\hat{\textbf{T}}$ as inputs to estimate the pose parameters $\hat{\theta}$, which parameterize the 3D rotations of the kinematic joints, enabling the animation of the human body with the desired poses.

\section{Prompts generated by ChatGPT}
We task ChatGPT to generate diverse characters (real-life celebrities and fictional) and diverse scene descriptions aligned with templates like "winter scene, snowy scene, beautiful snow." For each video, we randomly select over two character/scene combinations from the generated outputs as target prompts, shown in Tab. \ref{tab:example}.
\section{More visual and comparison results}
\begin{figure}[t]
\centering
\setlength{\abovecaptionskip}{0pt}
\includegraphics[width=1.0\textwidth,height=1.0\textheight]{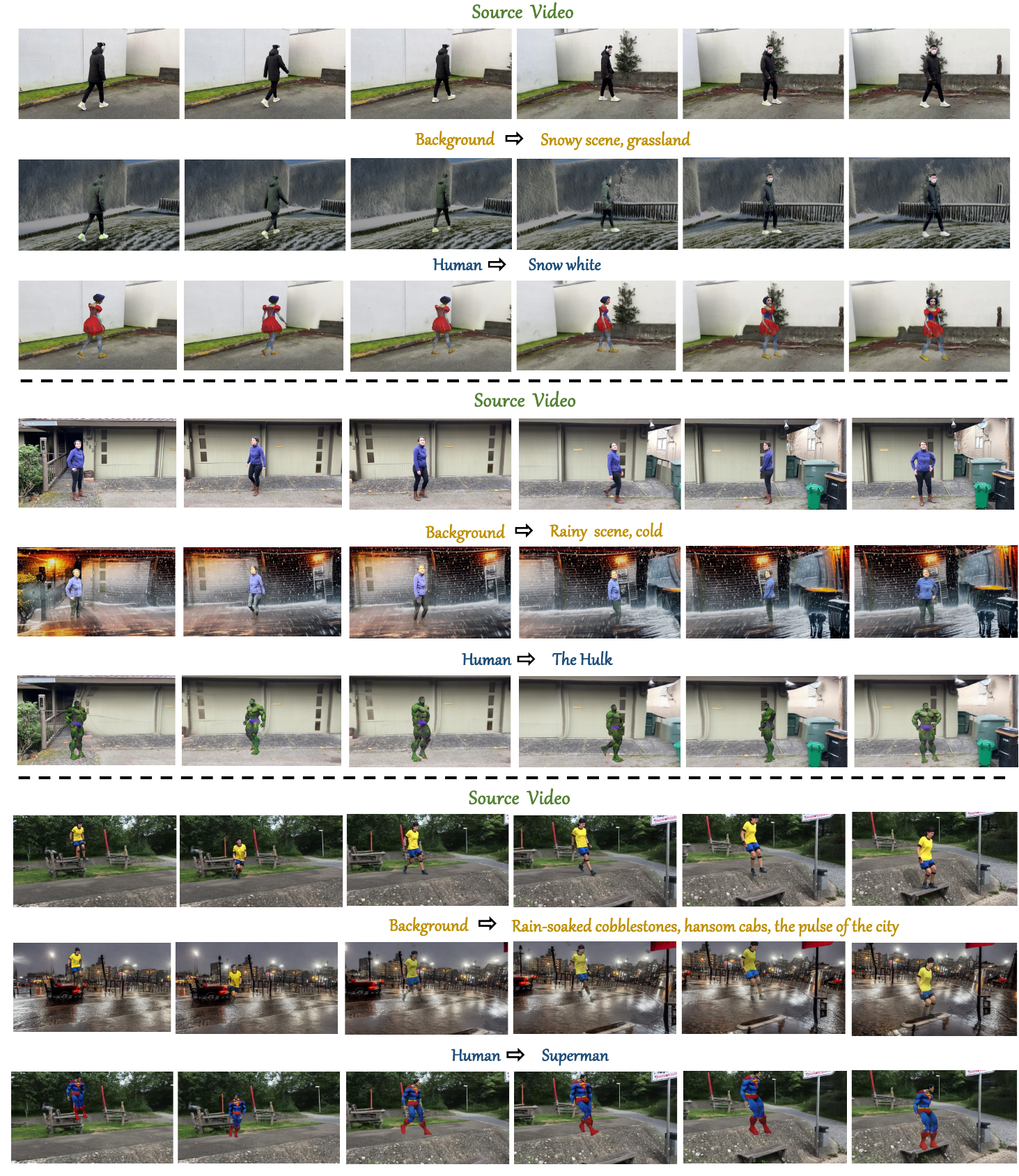}
\caption{\textbf{Visual results of editing humans or background separately.}}
\label{1}
\end{figure}

\begin{figure}[t]
\centering
\setlength{\abovecaptionskip}{0pt}
\includegraphics[width=1.0\textwidth,height=1.0\textheight]{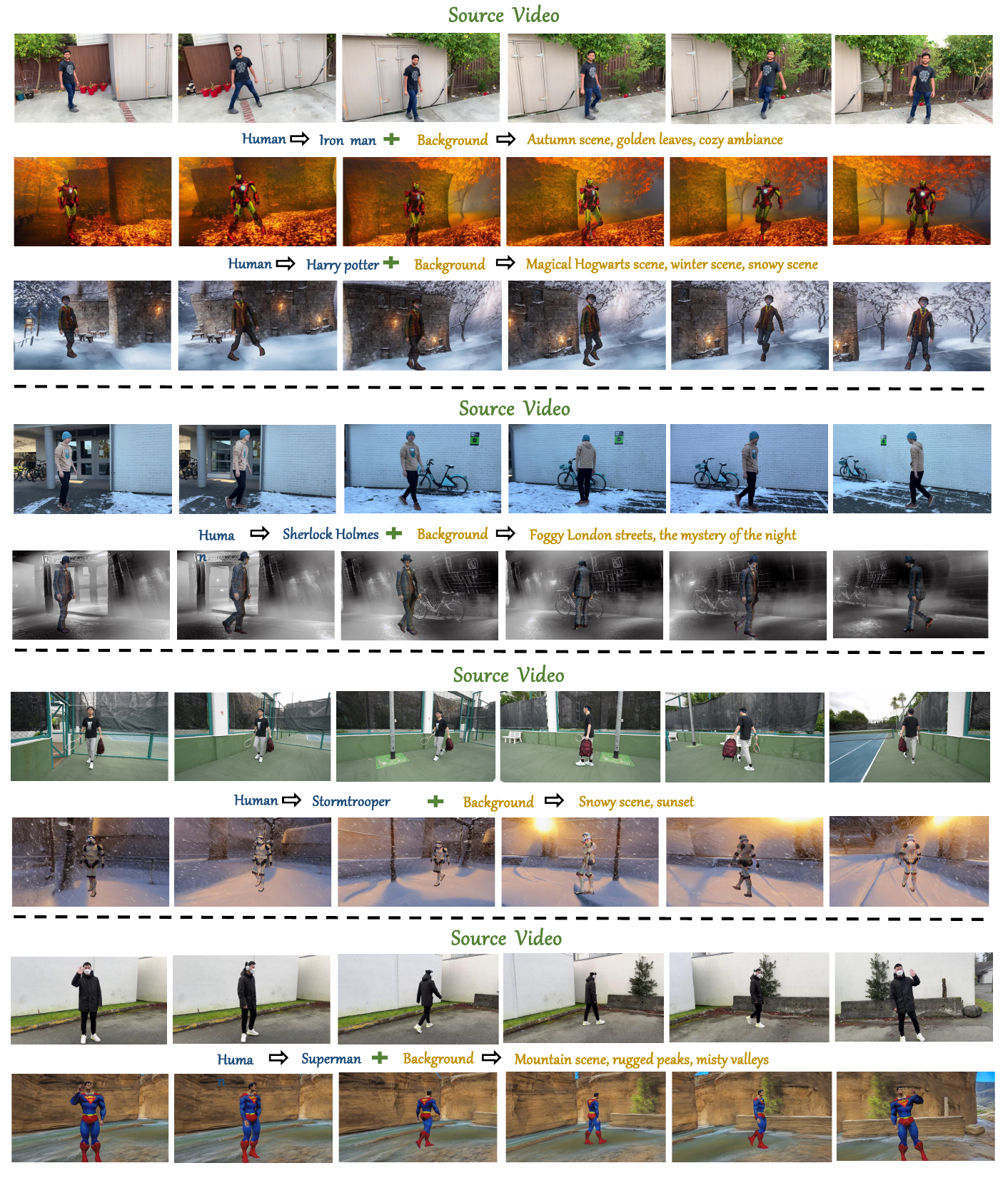}
\caption{\textbf{Visual results of editing humans and background simultaneously.}}
\label{2}
\end{figure}

\begin{figure}[t]
\centering
\setlength{\abovecaptionskip}{0pt}
\includegraphics[width=1.0\textwidth,height=1.0\textheight]{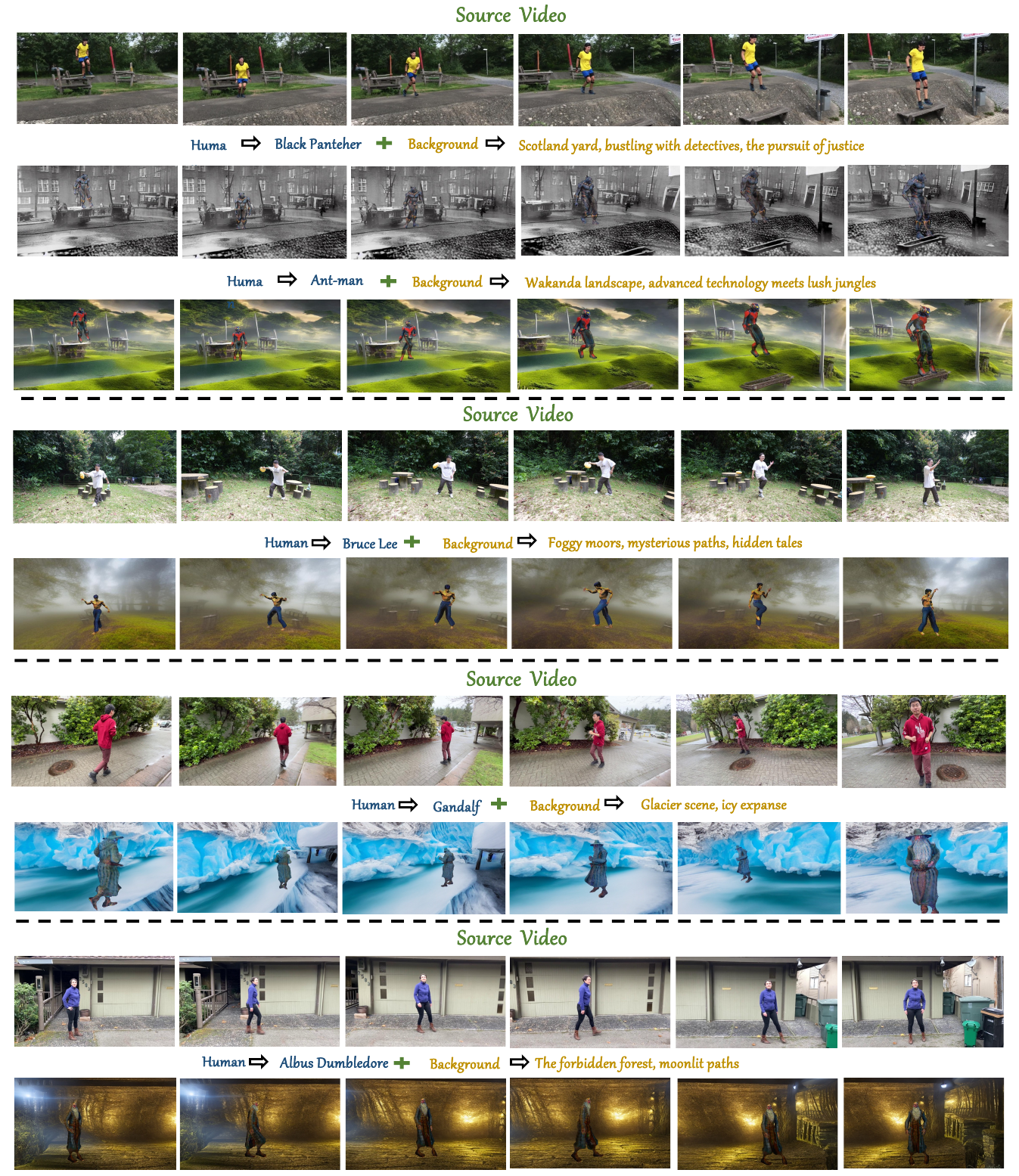}
\caption{\textbf{Visual results of editing humans and background simultaneously.}}
\label{3}
\end{figure}

\begin{figure}[t]
\centering
\setlength{\abovecaptionskip}{0pt}
\includegraphics[width=1.0\textwidth,height=1.0\textheight]{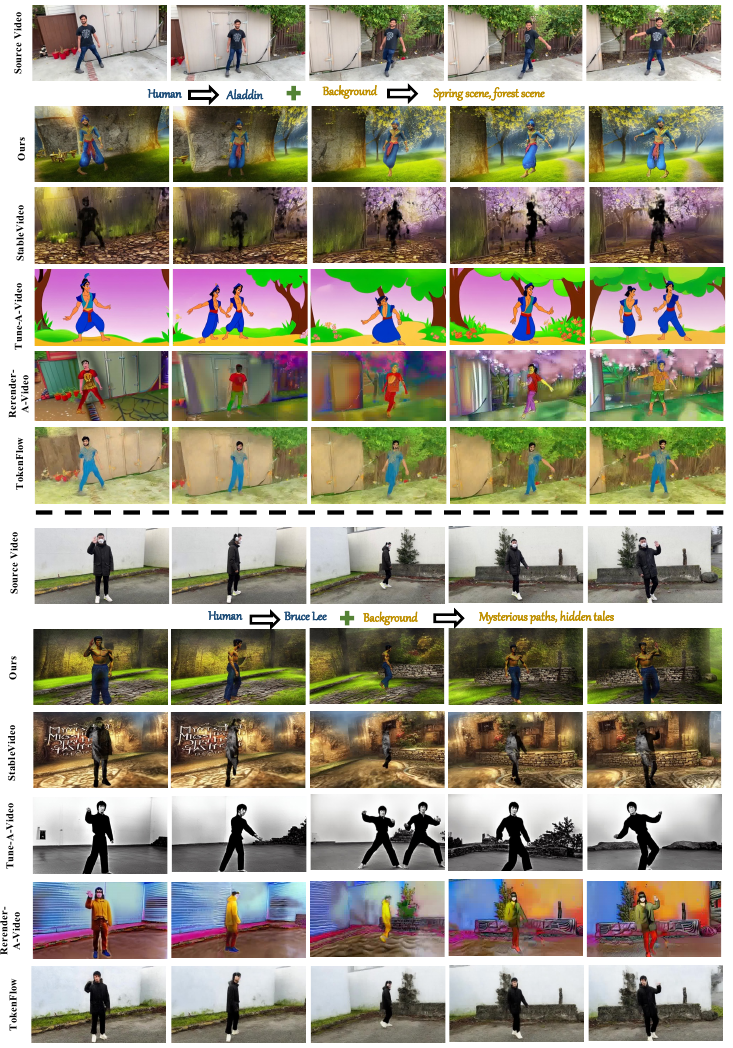}
\caption{\textbf{Qualitative comparisons of Deco and four representative video editing methods.}}
\label{4}
\end{figure}

\begin{figure}[t]
\centering
\setlength{\abovecaptionskip}{0pt}
\includegraphics[width=1.0\textwidth,height=1.0\textheight]{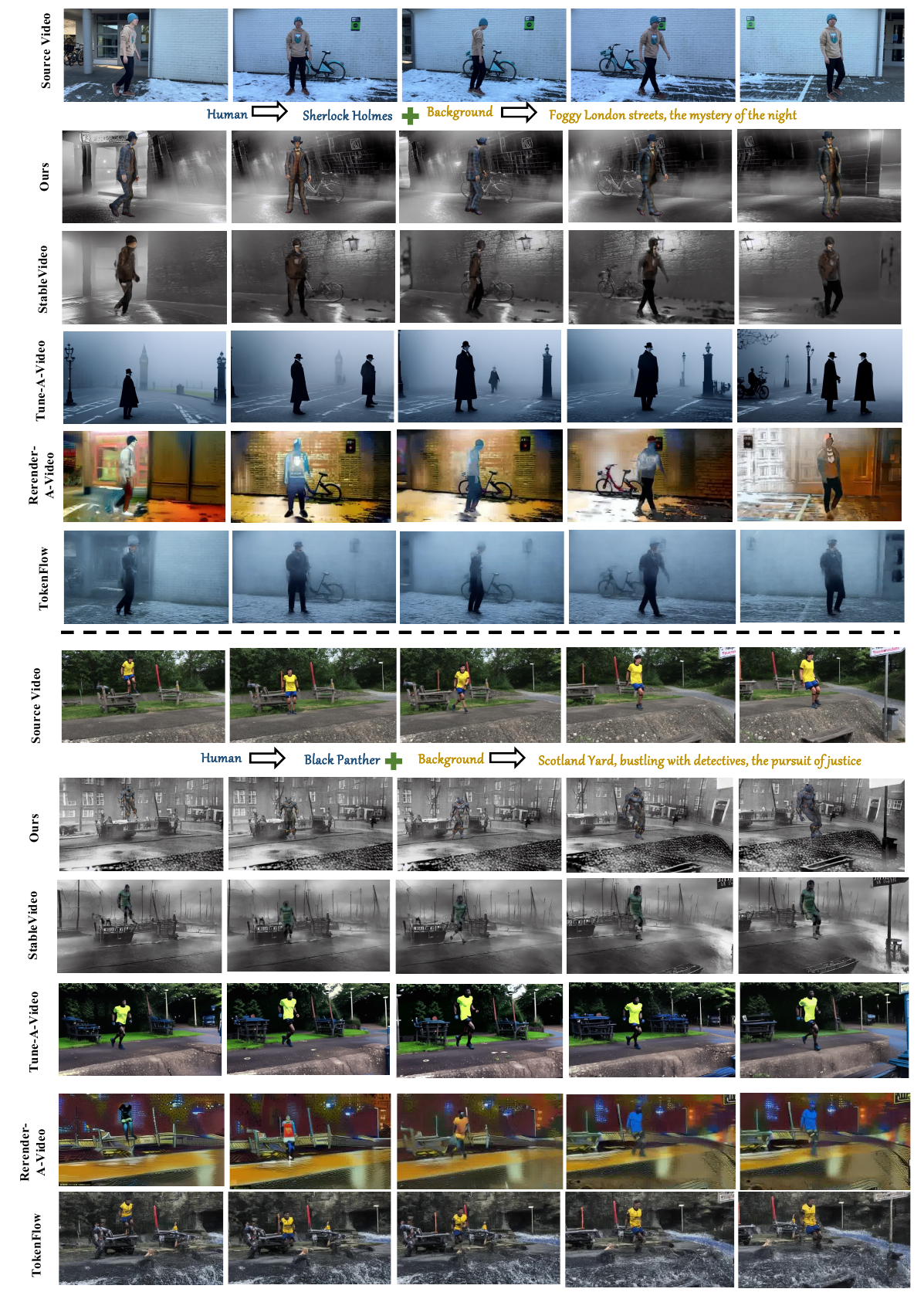}
\caption{\textbf{Qualitative comparisons of Deco and four representative video editing methods.}}
\label{5}
\end{figure}

As illustrated in Fig. \ref{1}, we provide the results of editing the human while preserving the original background, and editing the background while preserving the original human. Note that when we place the original human into a new background, the corresponding lighting changes accordingly. This is due to our proposed lighting-aware video harmonizer. Fig. \ref{2} and Fig. \ref{3} showcase the results of simultaneously editing both the human and the background. As can be observed, our approach allows for flexibly specifying the editing target, whether it is the human, the background, or both simultaneously. We provide more qualitative comparison results with StableVideo \cite{chai2023stablevideo}, Tune-A-Video \cite{wu2023tune}, Rerender-A-Video \cite{yang2023rerender} and TokenFlow \cite{geyer2023tokenflow} in Fig. \ref{4} and Fig. \ref{5}.

\bibliographystyle{splncs04}
\bibliography{egbib}

\begin{table}[htbp]
\centering 
\caption{Prompts generated by ChatGPT} 
\label{tab:example} 
\begin{tabular}{|c|c|}
\hline 
\textbf{Characters} & \textbf{Scenes} \\ \hline 
Mulan & Snow-capped mountains, clear blue sky, crisp air. \\
Stormtrooper & Ancient ruins, historical awe, whispers of the past. \\
Loki & Urban park, peaceful walks. \\
Gandalf & Glacier scene, icy expanse.\\
Sherlock Holmes & Foggy London streets, the mystery of the night. \\
Wonder Woman & Old town, cobblestone streets, echoes of history.\\
Ant-man & Rain-soaked cobblestones, hansom cabs, the pulse of the city.\\
Black Panther & Scotland Yard, bustling with detectives, the pursuit of justice. \\ 
Harry Potter & Hogwarts School of Witchcraft and Wizardry, \\
Superman & ancient stone walls echoing with magic. \\
Albus Dumbledore& The Forbidden Forest, moonlit paths.\\
The Hulk & Mountain scene, rugged peaks, misty valleys. \\
Snow White & Garden scene, blooming flowers, tranquil paths. \\
Batman & Winter scene, snowy scene, beautiful snow. \\
Spiderman & Dense jungle, mysterious paths, exotic sounds. \\
Aquaman & Underwater reef, coral gardens, swimming with fish. \\
Bruce Lee & Foggy moors, mysterious paths, hidden tales. \\
Iron Man & Battle scene, chaotic destruction. \\
Barack Obama & Countryside scene, rolling hills, quaint farmhouses.\\ 
Joker & Abandoned castle, gothic arches, stories untold. \\ 
Aladdin & Spring scene, forest scene. \\ 
Captain America & Summer scene, sunlit beach, gentle waves. \\ 
\hline 
\end{tabular}
\end{table}

